\documentclass{article}
\usepackage{subcaption}
\usepackage{float}
\usepackage{hyperref}
\usepackage{graphicx}
\usepackage{listings}
\usepackage{xcolor}
\definecolor{codegreen}{RGB}{0,255,0}
\definecolor{codewhite}{RGB}{128,128,128}
\definecolor{codepurple}{RGB}{200,0,200}
\definecolor{backcolour}{RGB}{255,255,255}
\usepackage[margin=1in]{geometry}
\lstdefinestyle{mystyle}{
    backgroundcolor=\color{backcolour},   
    commentstyle=\color{codegreen},
    keywordstyle=\color{magenta},
    numberstyle=\tiny\color{codewhite},
    stringstyle=\color{codepurple},
    basicstyle=\ttfamily\footnotesize,
    breakatwhitespace=false,         
    breaklines=true,                 
    captionpos=b,                    
    keepspaces=true,                 
    numbers=left,                    
    numbersep=5pt,                  
    showspaces=false,                
    showstringspaces=false,
    showtabs=false,                  
    tabsize=2
}
\begin{document}
\title{\rule{\textwidth}{3pt}\\\vspace{-0.7em}\rule{\textwidth}{2pt} PartitionVAE --- a human-interpretable VAE\\\rule{\textwidth}{2pt}\vspace{-0.7em}\\\rule{\textwidth}{3pt}}
\author{Fareed Sheriff\footnote{fareeds@mit.edu} \hspace{2em} Sameer Pai\footnote{sampai@mit.edu}\\\hspace{0.5em} MIT\hspace{6.0em} MIT}
\maketitle
\begin{abstract}
    \par VAEs, or variational autoencoders, are autoencoders that explicitly learn the distribution of the input image space rather than assuming no prior information about the distribution. This allows it to classify similar samples close to each other in the latent space's distribution. VAEs classically assume the latent space is normally distributed, though many distribution priors work, and they encode this assumption through a K-L divergence term in the loss function. While VAEs learn the distribution of the latent space and naturally make each dimension in the latent space as disjoint from the others as possible, they do not group together similar features --- the image space feature represented by one unit of the representation layer does not necessarily have high correlation with the feature represented by a neighboring unit of the representation layer. This makes it difficult to interpret VAEs since the representation layer is not structured in a way that is easy for humans to parse.
    \par We aim to make a more interpretable VAE by partitioning the representation layer into disjoint sets of units. Partitioning the representation layer into disjoint sets of interconnected units yields a prior that features of the input space to this new VAE, which we call a partition VAE or PVAE, are grouped together by correlation --- for example, if our image space were the space of all ping ping game images (a somewhat complex image space we use to test our architecture) then we would hope the partitions in the representation layer each learned some large feature of the image like the characteristics of the ping pong table or the characteristics and position of the players or the ball. We also add to the PVAE a cost-saving measure: subresolution. Because we do not have access to GPU training environments for long periods of time and Google Colab Pro costs money, we attempt to decrease the complexity of the PVAE by outputting an image with dimensions scaled down from the input image by a constant factor, thus forcing the model to output a smaller version of the image. We then increase the resolution to calculate loss and train by interpolating through neighboring pixels. We train a tuned PVAE on MNIST and Sports10 to test its effectiveness.
\end{abstract}
\section{Introduction}
\par Variational autoencoders \cite{VAE} are a type of neural network that attempts to learn the distribution of the input space while imposing the prior that the input space is normally distributed. The neural network (a multi-layer perceptron) is autoencoder-like in that the hidden layer learns the input space as each input is passed through the network. The prior that the hidden layer, the representation layer, is normally distributed is maintained by introducing a K-L divergence term in the loss function, which itself includes some measure of difference between the input and output images (mean squared error or binary cross-entropy, for example, depending on what is known about the values over which the input ranges). VAEs represent the latent space as a distribution by learning the mean and log variance of the latent space, which allows us to sample from the latent space to produce viable entries in the input space. While effective at encoding the input space in the usually more compact latent space, VAEs are classically difficult to interpret. Specifically, the latent space has no predefined structure beyond the prior that it is normally distributed.
\par Game representation \cite{Mandziuk2010} is a long-standing field of interest that focuses on finding efficient representations of games. The importance of this problem is visible in both the real world and the machine learning world: efficient game representations can increase the efficiency and potentially accuracy of intelligent game agents, and compact game representations are often easier for humans to understand than more dense representations. Furthermore, efficient game representations can be used in conjunction with common game representations to simplify common representations without losing too much information, which is a good trade-off for both humans and machines. 
\par Various architectures have been designed to increase the interpretability of the VAE, discussed in more detail in the next section. The PVAE seeks to increase interpretability by grouping together similar features (features of the input space with high correlation). The end result is a partitioned representation layer each of whose partitions contains a set of correlated units. The representation layer is partitioned into groups of neurons that are each connected to a neural network whose output is of the same size as the size of each group of neurons. This puts a large portion of the PVAE into disjoint sets of neurons that together form the representation, which pushes the representation to be partitioned such that individual partitions of the representations represent features of the input space that correlate with each other and features that do not correlate with each other that much are generally not represented by neurons in different partitions. The partition feature of the PVAE is especially useful for game representations because based on the number of partitions, we can isolate individual important components of an image representation of a game. In ping pong, for example, a given partition could encode the characteristics of the ping pong table.
\section{Related Work}
\par Significant research has been done both on game representations and making VAEs more interpretable. Some examples of more interpretable VAEs are oi-VAE (output interpretable VAE) \cite{pmlr-v80-ainsworth18a} and PI-VAE (physics-informed VAE) \cite{PIVAE}. oi-VAEs attempt to disentangle latent variables by penalizing for multiple variables sharing data. This attempts to keep units in the representation layer as disjoint as possible in terms of what they represent. In general terms, PI-VAE attempts to make the method of training interpretable by modeling the VAE through physics processes (by training on differential equation data using a physics-based loss functions), making the model a subset of a field of ML known as physics-based ML. While both seek to make the VAE more interpretable in different ways, it should be noted that oi-VAE, which is closest to our definition of interpretability (making the representation layer more interpretable), acts specifically on latent variables by decreasing correlation between variables. In contrast, PVAE encourages information-sharing between variables that already have high correlation and prevents interaction between latent variables within different partitions and therefore lower correlation. PVAE therefore aims to be more interpretable over partitions rather than over latent variables themselves.
\par Various articles have been published on general game representation, including a paper on creating representations of video games that are edition- and graphics-invariant \cite{conmodel}, an article on contrastive learning between games based on genre that aimed to learn representations of games by genre rather than specifically by game \cite{gengame}. Furthermore, play2vec, a sports representation robust to noise, is a sports representation model that attempts to compare sports plays by comparing the similarity of their representations \cite{dlsportsretrieval}. While each of these models learn representations that are invariant over some category, they do not explicitly create a game representation optimized for individual games. Because they are supposed to learn invariant representations, efficient representation is not the main goal of these papers but more a secondary goal if at all. We apply PVAE to images of ping pong matches to evaluate our primary goal of creating a compact and accurate but interpretable representation of the ping pong game image space, contrasting in purpose with the models mentioned above.
\section{Methods}
\par We describe how we define and test our PVAE in this section. PVAE is implemented in Python using the PyTorch library. We test PVAE on MNIST and Sports10 with a variety of parameters to analyze how well the representation encodes samples. Finally, we examine sample representations and examine the effects of modifying the representation partition-by-partition.
\par A PVAE is a VAE with three parts: an encoder, a partition, and a decoder. The encoder consists of a set of convolution layers as defined by the user followed by a Flatten layer and a Linear layer to share a small bit of information between units and to allow partitions in the representation to assign information to partitions based on the information's dimensionality. The partition layer consists of ANNs that preserve partition size, thus learning features of the representation, two sets for each partition to yield the mean and log variance. We sample from the mean and log variance layers and pass this through the decoder. Finally, the decoder consists of Linear layers and alternating conv and deconv layers to rebuild an output image from the representation. The loss function is a weighted sum of reconstruction loss and K-L divergence, the former to ensure the output resembles the input and the latter to maintain the prior that the representation is normally distributed. We weight the reconstruction loss (MSE loss) multiple magnitudes more than the K-L divergence when testing PVAE because the normal prior is far stronger than the reconstruction prior. We test representations of size $k$ on MNIST and Sports10 by comparing the losses of different partitions of $k$ with how badly the model overfits.
\par Because we do not have free access to GPUs, we take measures to decrease the complexity of training our PVAE. We do this by making the PVAE yield an output image smaller than the input image, then upscale the output to the resolution of the input. In this way, the representation does not need to be as exact and the decoder is less complex. Finally, we slightly perturb the train data by inserting low levels of random noise into training samples. Examples of different components of a sample PVAE are displayed in the appendix.
\section{Results}
\subsection{Datasets} 

\par In this section we describe the datasets that we use to train and evaluate our model. The first dataset used was the MNIST hand-written digits dataset, which is commonly used for computer vision recognition tasks. We chose MNIST as it is a relatively small dataset both in terms of number of images and resolution, making it easy to quickly iterate and test our model. 

\par For a more complex dataset on which we can test our PVAE model, we use the Sports10 dataset, a set of over 100K images of sports video games first introduced in \cite{gengame}. The dataset is split into ten different sports, such as football, fighting, and table tennis. For our project we chose to use only the table tennis subset of the images. Furthermore, due to computational constraints, we split this collection of table tennis images into a smaller set of roughly 1K images, which we divided into 900 training images and 100 validation images randomly.

\begin{figure}
    \centering
    \begin{subfigure}[b]{0.4\textwidth}
    \includegraphics[width=\textwidth]{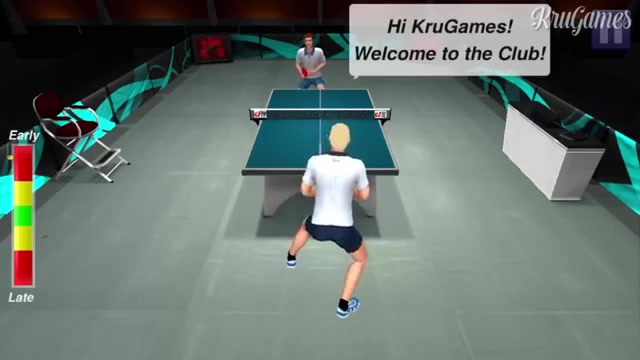}
    \end{subfigure}
    \begin{subfigure}[b]{0.4\textwidth}
    \includegraphics[width=\textwidth]{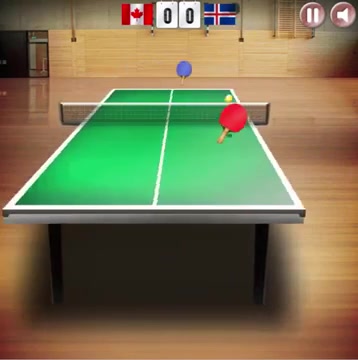}
    \end{subfigure}
    \caption{Examples of images from the Table Tennis dataset.}
\end{figure}
\subsection{Results on MNIST Dataset}
\begin{figure}[h]
    \centering
    \begin{subfigure}[b]{0.4\textwidth}
    \includegraphics[width=\textwidth]{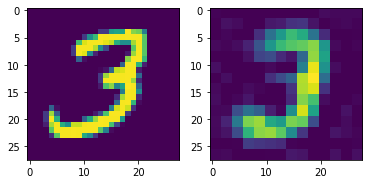}
    \end{subfigure}
    \begin{subfigure}[b]{0.4\textwidth}
    \includegraphics[width=\textwidth]{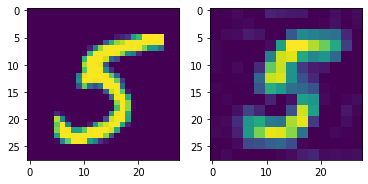}
    \end{subfigure}
    \begin{subfigure}[b]{0.4\textwidth}
    \includegraphics[width=\textwidth]{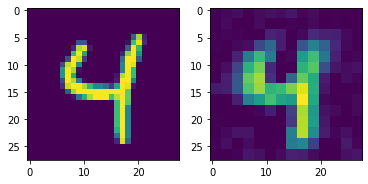}
    \end{subfigure}
    \caption{Examples of upsampled PVAE output after training on the MNIST dataset}
    \label{mnist_reconstruction}
\end{figure}

We trained a PVAE model on the MNIST dataset, with a representation size of 10 (split into partitions of sizes 4, 3, and 3). A downsampling factor of $2$ was used, so that the model would output images at half the resolution of the input, and then upsampled to compute loss. After 25 epochs of training, the model was able to reliably output images that were visually similar to the input images (see Figure \ref{mnist_reconstruction} for sample reconstructions of images). The output images of the model are blurrier than the input because, as described earlier, the model outputs an image that is half the resolution of the input image, and then upsampled. 

\begin{figure}[h]
    \centering
    \begin{subfigure}[b]{0.4\textwidth}
    \includegraphics[width=\textwidth]{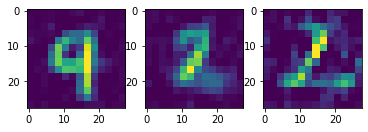}
    \caption{Scaling down the first partition (size 4) in the representation unravels the 9 and adds a ``tail,'' changing it into a 2.}
    \end{subfigure}
    \begin{subfigure}[b]{0.4\textwidth}
    \includegraphics[width=\textwidth]{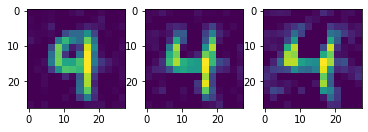}
    \caption{Scaling down the second partition (size 3) separates the top of the 9, making it look like a 4.}
    \end{subfigure}
    \begin{subfigure}[b]{0.4\textwidth}
    \includegraphics[width=\textwidth]{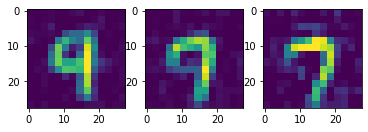}
    \caption{Scaling down the third partition (size 3) unravels the 9 but doesn't add a tail, making it a 7.}
    \end{subfigure}
    \caption{Effects of scaling individual partitions on PVAE output, on a sample image of a 9.}
    
    \label{mnist_features}
\end{figure}

Now that we have learned a low-dimensional representation (of size 10), we can investigate the effect that changing different elements of the representation has on the decoded output. Specifically, we will take all the elements corresponding to a single partition, and scale only those elements in the representation. Figure \ref{mnist_features} shows this process, applied to a sample test image of a $9$. We observe that scaling each of the partitions corresponds to an easily describable transformation to the number, such as unfolding the top to make it a 4. 

Furthermore, we can see that each of the three partitions transform the digit into three separate classes, providing evidence that the partitions operate independently of each other. Qualitatively, our PVAE has proved to be quite successful on MNIST: it provides a representation from which we can accurately reconstruct the original digit image, and furthermore the partitions in the representation interact with the output image in a way that is readily interpretable by a human observer. 

Now that the PVAE has shown promising results on MNIST, we are ready to evaluate it on our (significantly more complex) target dataset: the Table Tennis section of Sports10.


\subsection{Results on Table Tennis Dataset}

\begin{figure}[h]
    \centering
    \begin{subfigure}[b]{0.4\textwidth}
    \includegraphics[width=\textwidth]{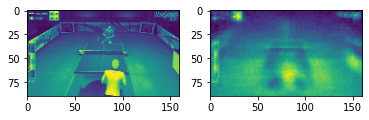}
    \end{subfigure}
    \begin{subfigure}[b]{0.4\textwidth}
    \includegraphics[width=\textwidth]{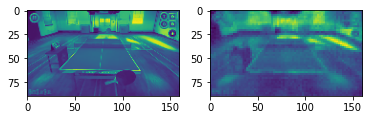}
    \end{subfigure}
    \begin{subfigure}[b]{0.4\textwidth}
    \includegraphics[width=\textwidth]{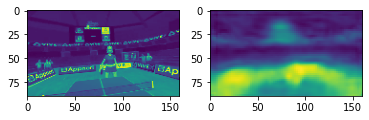}
    \end{subfigure}
    \caption{Examples of upsampled PVAE output after training on the Table Tennis dataset}
    \label{tt_reconstruction}
\end{figure}

After randomly generating train and test splits of the Table Tennis dataset, we trained our PVAE with a representation size of 20 (split into partition sizes of 5, 5, 4, 3, 2, and 1). We trained the model for 50 epochs, after which both training and validation loss plateaued. Figure \ref{tt_reconstruction} shows some example output of the PVAE. In general, the model seems to be capable of capturing static aspects of the game such as the table, net, and surroundings, but elements that frequently change location are either absent from the output (in the case of the paddle on the lower image) or are displayed in a ``default'' location (the player in the top image is displayed at the center despite the input image having him on the right side). However, the model is clearly able to differentiate between the settings of the different games in the dataset, as evidenced by the camera angle and surroundings of the output. 

\begin{figure}[h]
    \centering
    \begin{subfigure}[b]{0.4\textwidth}
    \includegraphics[width=\textwidth]{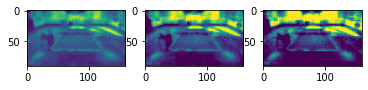}
    \caption{The first partition of size 5. Scaling this up greatly increases contrast in the image, while keeping other elements relatively stable. This partition could capture variation in contrast between different games within the dataset.}
    \end{subfigure}
    \begin{subfigure}[b]{0.4\textwidth}
    \includegraphics[width=\textwidth]{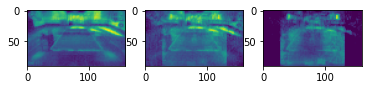}
    \caption{The second partition of size 5. This appears to impact the letterboxing of the image, with higher values introducing a horizontal margin on the output. The Table Tennis dataset contains images from both widescreen and square resolution games, which is why some input images have similar letterboxing.}
    \end{subfigure}
    \begin{subfigure}[b]{0.4\textwidth}
    \includegraphics[width=\textwidth]{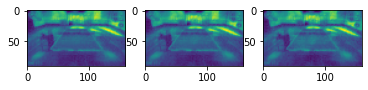}
    \caption{On the other hand, scaling on the third partition seems to have little to no effect on the output image. This shows a potential shortcoming in the PVAE model, since ideally every element of the representation should have an impact on the output image, especially if the PVAE is unable to fully reconstruct the input image.}
    \end{subfigure}
    \caption{Effects of individual partitions on the model output.}
    \label{tt_features}
\end{figure}

Similarly to the MNIST dataset, we can investigate how various elements of the representation impact the output image. Since the representation is split into partitions, we start by looking at each partition individually. Figure \ref{tt_features} shows some effects of various partitions of the learned representation on the PVAE output. Some of the partitions have very tangible effects on the output, affecting parts of the image such as the contrast (which varies greatly between games in this genre), as well as the letterboxing on the image (which is important due to differing screen resolutions between games). Qualitatively, the low-dimensional representation of the game image that is learned by the PVAE does capture some important differences between games in the table tennis genre.

However, there are some shortcomings of the PVAE models learned from this dataset. First of all, the output image is blurry and often omits details such as the player and opponent position. Especially when the input image is at a very oblique angle to the table, the model output is unrecognizable. This is due in part to the fact that we downscale the input image, and upsample at the end. We made this decision in order to speed up the training time and reduce the reliance on compute power. Indeed, when we tried removing the downsampling, the model took twice as long to train and still was unable to achieve similar results to the original (downsampled) model.

Another issue is that some of the partitions in the representation seem to have little to no tangible effect on the output image, demonstrating that we are not learning an efficient representation on the input dataset. Consider, for example, the third partition in Figure \ref{tt_features}. This partition does not seem to affect the output image in any meaningful way, even with large amounts of scaling. 

Many variations on the hyperparameters and dataset were attempted to achieve better results. Some of the experiments carried out include:
\begin{itemize}
    \item Removing the downsampling factor, so that the output and input image match in resolution.
    \item Increasing and decreasing the representation size
    \item Training the PVAE on only one game, to see if a better representation can be learned on a simpler dataset
\end{itemize}

\begin{figure}
    \centering
    \includegraphics[width=0.5\textwidth]{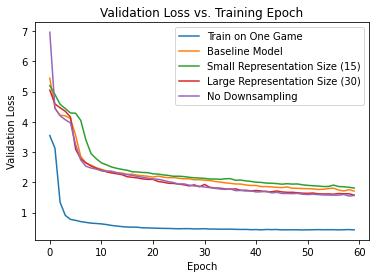}
    \caption{Loss curve for different variations on PVAE hyperparameters and dataset. With the exception of the reduced dataset, all variations exhibit similar loss curves.}
    \label{ttloss}
\end{figure}

Quantitatively, the simplest method to evaluate the success of all of these variations on the PVAE is by measuring the combined K-L and MSE losses on the models, since this is the loss function that we train the auto-encoder to minimize. Figure \ref{ttloss} shows the loss of each of the variations above as a function of the number of training epochs. We can see that many of the variants do achieve better loss than the original model.
However, they did not bring us closer to our original goal, which was to develop a learned representation that is more easily human-interpretable. In most cases, the PVAE still outputs a blurry image that seems to average out most of the individual differences that happen within a game image. Consider, for example, the game in Figure \ref{tt_features}, which has a relatively fixed perspective and background. The PVAE can reliably reconstruct the static table and background, but not the paddle. On games with highly dynamic perspectives, such as the last one in Figure \ref{tt_reconstruction}, the PVAE fails at reconstruction.
Therefore, we can conclude that, while our PVAE model is capable of learning various interesting features on this dataset, ultimately more work is needed to develop it into a model that can effectively learn an interpretable representation on a complex dataset.
\section{Conclusion}
\par In conclusion, we have developed a unique variant of an auto-encoder that splits the representation layer into partitions, with the goal of encouraging information-sharing within a partition and discourage sharing between partitions. It was our belief that in doing this, we would be able to generate an efficient but interpretable representation of an image dataset.  Through our experiments, we have discovered that our PVAE model is successful in some areas, yet falls short in others. On a simple dataset such as MNIST, our models are capable of faithfully reconstructing the input, and furthermore each partition was found to perform comprehensible transformations on a sample input image. This is desirable behavior for our PVAEs. However, our models had more mixed results on the Table Tennis dataset. Specifically, on the qualitative level, our models have proven to be quite capable of distinguishing between different games within the table tennis genre, and captures static aspects of these games such as the table position, surroundings, and contrast quite well. Furthermore, these characteristics can be seen within the partitions of the representations, with image contrast and screen resolution being directly controlled by two different partitions in our trained model. Quantitatively, our experiments with varying representation sizes and simpler datasets show that our model converges relatively quickly. However, we have been unable to tune a PVAE to accurately reconstruct more dynamic details in a game image, such as the positions of humans or the viewing angle. Furthermore, regardless of how we configured the downsampling or model architecture, the model was unable to learn an efficient representation, since a large portion of the partitions either did not seem to have a major effect on the output image, or affected too much of the output to be easily interpretable. This is likely because scenes as complex as table tennis games have small but relevant differences between individual frames to be efficiently compressed by a VAE without significant preprocessing, something that might be able to be mitigated through longer training. Another possible explanation for the shortcomings of this model is the fact that these games actually have very few moving parts. Therefore, it is easy for the PVAE to average out differences within a class and achieve low MSE error. If we want our model to be able to represent intra-class differences, one possible idea would be to structure our loss function in a way that better penalizes stationary outputs over a single class. Another concern that we had about our model was that the encoder and decoder architecture were simply not robust enough to represent a complex and varied dataset, mainly because VAEs on the whole are not a general-purpose model you can simply throw onto any CV problem. Further investigation into different model architectures could be a viable avenue for exploration.
\section{Further Research}
\par Our training and results have led us to consider various directions for further research, including comparing PVAE representation sizes to vanilla VAE representation sizes at equivalent levels of loss/image recognizability, increasing model complexity, increasing input size, and training for longer/finetuning training. We noted that in our preliminary results, PVAE trained on MNIST worked surprisingly well at small representation sizes and provided results similar to VAE on MNIST at larger representation sizes. This could be because the prior imposed on PVAE to group together correlated information means that less latent variables represent the same information, naturally decreasing the representation size. One further area of research is using PVAE to yield smaller representations without signficantly impacting reconstruction quality.
\par We also note that the input image in our tests were very small as large images would take an inordinate amount of time to train. As a result of using small images, details in the image were likely lost and thus reconstruction was understandably difficult. However, increasing the image size to that of a normal image would almost certainly improve model performance though the model might take significantly longer to converge since the complexity of the image space increases with size.
\par Finally, we consider training for many times longer than we did when training on MNIST and Sports10 as we only trained for around 30 to 60 epochs on each dataset (a limitation of the lack of access to strong GPUs and time we had to train), but the loss notably never converged and instead kept decreasing over training, albeit at a decreasing rate of decrease. Also, the test loss was always some margin below the training loss, implying that the model was not overfitting and therefore still had a decent bit of training to do.
\newpage
\section{Appendix}
\appendix
\section{Model Diagrams}
\begin{figure}[H]
    \centering
    \includegraphics[width=8cm]{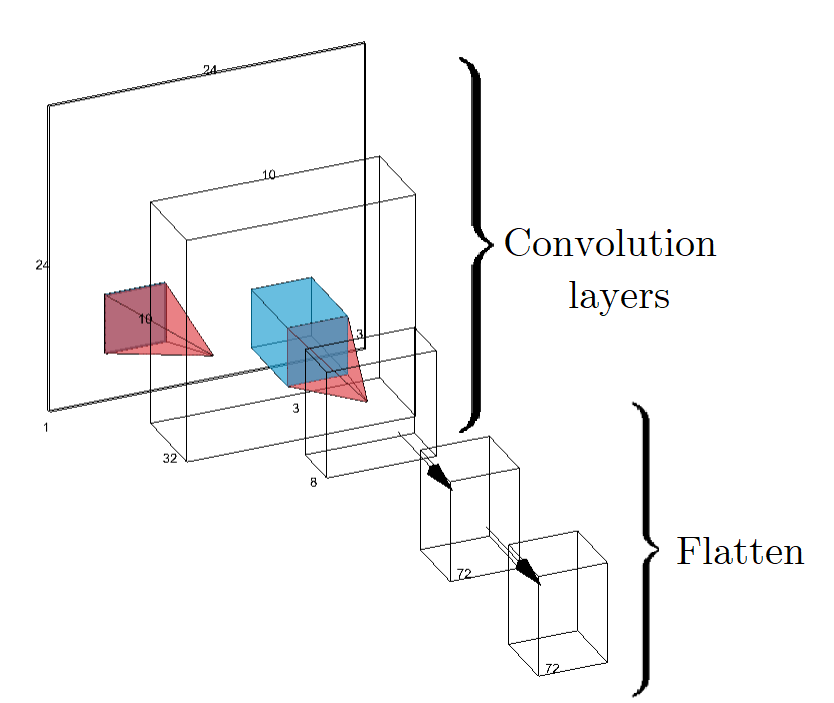}
    \caption{Sample PVAE encoder for MNIST}
\end{figure}
\begin{figure}[H]
    \centering
    \includegraphics[width=8cm]{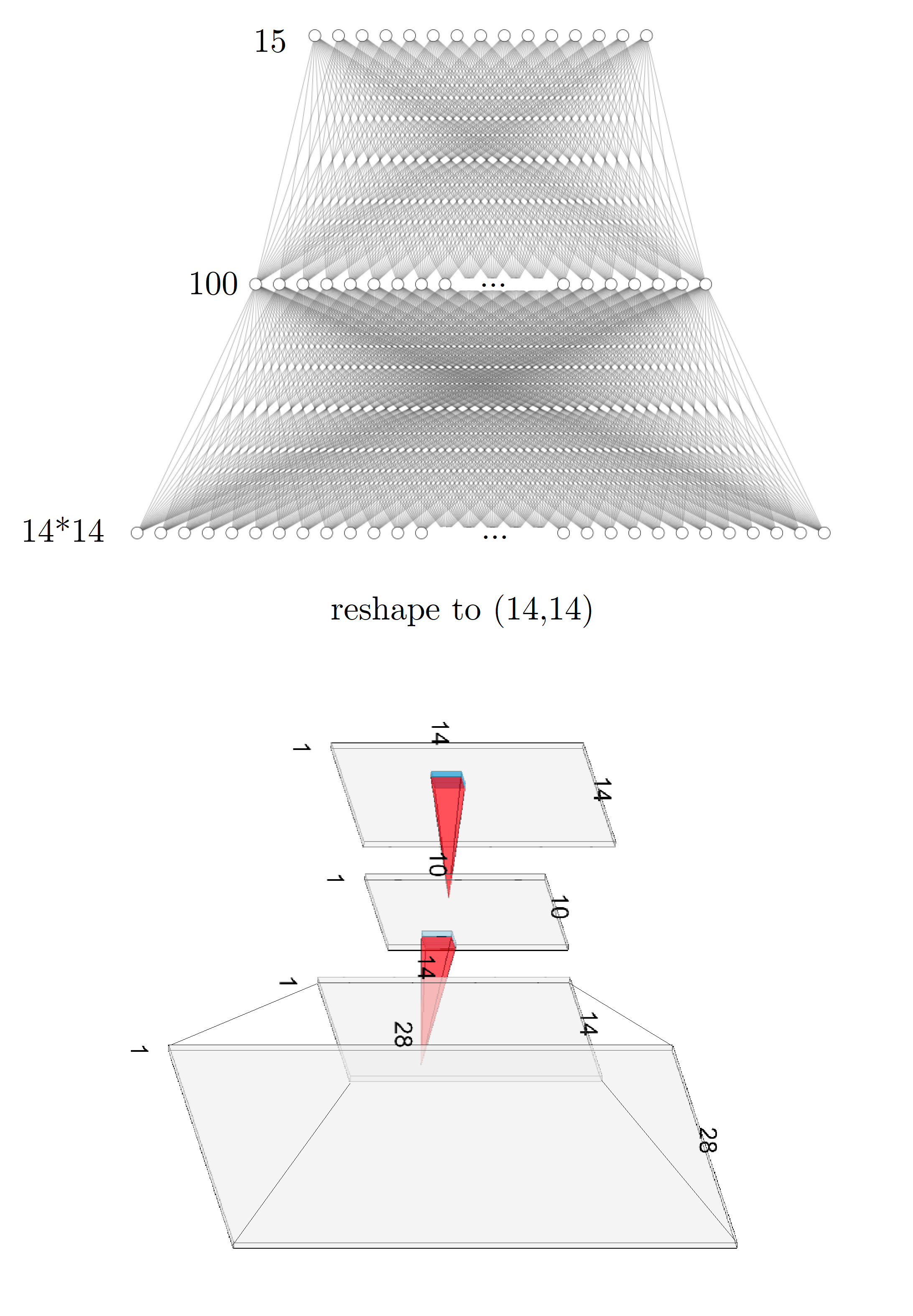}
    \caption{Sample PVAE decoder for MNIST}
\end{figure}
\begin{figure}[H]
    \centering
    \includegraphics[width=10cm]{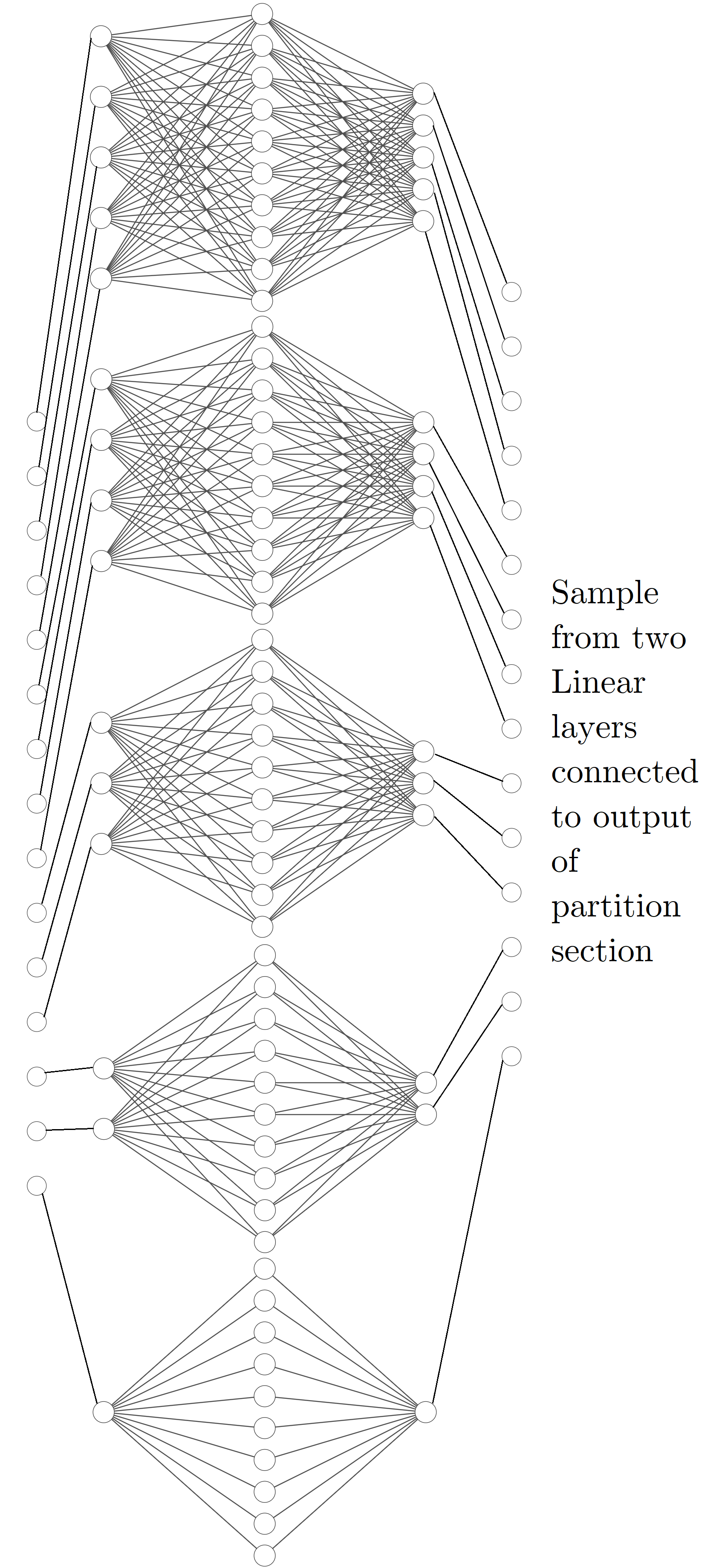}
    \caption{Sample partition layers for MNIST}
\end{figure}

\section{Model Code}
The code we wrote for the PVAE model is below. Complete code including the tests we wrote and ran as well as the utility functions we used can be found at \url{https://github.com/meeeeee/PartitionVAE}.
\begin{lstlisting}[language=Python]
import numpy as np
import torch
from torchvision.transforms import ToTensor
from torch.utils.data import DataLoader
import torch.nn as nn
import PIL
from tqdm import tqdm

class PVAE(nn.Module):
  """
  encoder, convlistdec: list of conv layers as lists/tuples with first entry kernel and second entry channels
  linlistdec: list of linear layer sizes
  partition: partition of representation layer as list of lists of Linear layer sizes
  """
  def __init__(self, encoder=[[5,32],[5,8]], partition=[[5,2],[10,4]], linlistdec=[100,100], convlistdec=[[5,1]]):
    super(PVAE, self).__init__()
    encoderlist,decoderlinlist,decoderconvlist = [],[],[]
    convshape, channels = np.array([f*h,f*w]), 1
    for auto in encoder:
      encoderlist.append(nn.Conv2d(in_channels=channels,out_channels=auto[1],kernel_size=auto[0],stride=1,padding='valid').to(device))
      encoderlist.append(nn.ReLU().to(device))
      encoderlist.append(nn.MaxPool2d(kernel_size=2).to(device))
      channels = auto[1]
      convshape = ((convshape-auto[0]+1)/2).astype(int)
    encoderlist.append(nn.Flatten().to(device))
    length = convshape[0]*convshape[1]*channels

    if 0 in [sum(auto) for auto in partition]: raise Exception("empty layer description in partition")
    self.partitions_mean, self.partitions_logvar = [], []
    self.split = []
    for auto in partition:
      lst = []
      partlen = auto[-1]#length
      self.split.append(auto[-1])
      for x in auto:
        lst.append(nn.Linear(in_features=partlen,out_features=x).to(device))
        partlen = x
      lst.append(nn.Linear(in_features=partlen,out_features=partlen).to(device))
      self.partitions_mean.append(nn.Sequential(*lst).to(device))
      self.partitions_logvar.append(nn.Sequential(*lst).to(device))
    
    encoderlist.append(nn.Linear(in_features=length,out_features=sum(self.split)).to(device))
    self.encoder = nn.Sequential(*encoderlist).to(device)

    length = sum([auto[-1] for auto in partition])
    
    if linlistdec[-1] != h*w: raise Exception("last layer of linlistdec does not equal h*w")
    for auto in linlistdec:
      decoderlinlist.append(nn.Linear(in_features=length, out_features=auto).to(device))
      length=auto
    self.lineardecoder = nn.Sequential(*decoderlinlist).to(device)

    convshape,channels = np.array([h,w]), 1
    for auto in convlistdec:
      decoderconvlist.append(nn.ConvTranspose2d(in_channels=channels,out_channels=auto[1],kernel_size=auto[0],stride=1,padding=0).to(device))
      decoderconvlist.append(nn.ReLU().to(device))
      channels = auto[1]
      decoderconvlist.append(nn.Conv2d(in_channels=channels,out_channels=channels, kernel_size=auto[0],stride=1,padding='valid').to(device))
      decoderconvlist.append(nn.ReLU().to(device))
    decoderconvlist.append(nn.Conv2d(in_channels=channels,out_channels=1,kernel_size=1,stride=1,padding='valid').to(device))
    self.convdecoder = nn.Sequential(*decoderconvlist).to(device)

  def normalsample(self, mean, logvar):
    std = torch.exp(logvar/2).to(device)
    eps = torch.randn_like(std).to(device)
    return mean+std*eps

  def forward(self, x):
    x = self.encoder(x) # encoder
    x = torch.split(x, self.split, dim=-1)
    xmean,xlogvar = [self.partitions_mean[a](x[a]) for a in range(len(self.split))],[self.partitions_logvar[a](x[a]) for a in range(len(self.split))] # pass partitioned units through Linear layers
    xmean,xlogvar = torch.cat(xmean,-1),torch.cat(xlogvar,-1) # merge partitioned units --- prerepresentation layer
    x = self.normalsample(xmean,xlogvar) # sample normally given mean and logvar
    x = self.lineardecoder(x) # linear layers
    x = x.view([x.shape[0],1,h,w]) # reshape for deconv
    x = self.convdecoder(x) # deconv layers, should maintain shape
    x = nn.functional.interpolate(x,scale_factor=f,mode='nearest-exact') # upsample to yield same size as input image
    return x,xmean,xlogvar # return output image and distribution parameters

  def __getitem__(self, args):
    mean,logvar = args
    y = self.normalsample(mean,logvar)
    y = self.lineardecoder(y)
    y = y.view([y.shape[0],1,h,w])
    y = self.convdecoder(y)
    y = nn.functional.interpolate(y,scale_factor=f,mode='nearest-exact')
    return y
\end{lstlisting}
\newpage
\bibliographystyle{unsrt}
\bibliography{refs.bib}
\end{document}